\def\year{2018}\relax
\documentclass[letterpaper]{article} 
\usepackage{aaai18}  
\usepackage{times}  
\usepackage{helvet}  
\usepackage{courier}  
\usepackage{url}  
\usepackage{multirow}
\usepackage{graphicx}  
\newcommand{\specialcell}[2][c]{%
  \begin{tabular}[#1]{@{}c@{}}#2\end{tabular}}
\begin{document}
%
\title{Automated Game Design via Conceptual Expansion}
\author{Matthew Guzdial and Mark Riedl\\
\\
School of Interactive Computing\\
Georgia Institute of Technology\\
Atlanta, GA 30332 USA\\
mguzdial3@gatech.edu, riedl@cc.gatech.edu\\
}
\maketitle
\begin{abstract}
Automated game design has remained a key challenge within the field of Game AI. 
In this paper, we introduce a method for recombining existing games to create new games through a process called conceptual expansion.
Prior automated game design approaches have relied on hand-authored or crowd-sourced knowledge, which limits the scope and applications of such systems. 
Our approach instead relies on machine learning to learn approximate representations of games. 
Our approach recombines knowledge from these learned representations to create new games via conceptual expansion.
We evaluate this approach by demonstrating the ability for the system to recreate existing games. 
To the best of our knowledge, this represents the first machine learning-based automated game design system.
\end{abstract}

\section{Introduction}

Game design and development requires a large amount of expert knowledge, in terms of design and coding skills. 
This skill requirement serves as a barrier that restricts those who might most benefit from the ability to make games. 
Researchers have touted automated game design as a potential solution to this issue, in which computational systems build games without major human intervention.
The promise of automated game design could not only democratize game design, but allow for educational, scientific, and entertainment applications of games currently infeasible given the resource requirements of modern game development.
However, up to this point automated game design has relied upon encoding human design knowledge in terms of authoring parameterized game design spaces or entire games for a system to remix.
This authoring work requires expert knowledge, and is time intensive to debug, which limits the applications of these automated game design approaches.

Procedural content generation (PCG), the automatic generation of content for games, represents a subset of the problem of automated game design and parallels it in terms of a burden of human authoring. PCG via Machine Learning (PCGML) instead trains models on existing game content to generate new game content and has been proposed as a solution to the design burdens of PCG \cite{summerville2017procedural}, 
For example, training a system on representations of Super Mario Bros. (SMB) levels to learn a distribution over levels and then sample from this distribution. 
However, PCGML requires a relatively large amount of existing content to build a distribution. 
Further, these methods require training data input that resembles the desired output.
For example, one couldn't train a machine learning system on SMB levels and expect to get anything but SMB-like levels as output.
That is, PGCML by itself is incapable of significant novelty.

In this paper we introduce a technique that recombines representations of games to produce novel games: {\em conceptual expansion}, a combinational creativity technique. This allows us to create novel games that contain characteristics of multiple existing games, based on machine learned representations of these games.
In this way, our approach can produce novel output in comparison to its input.
To the best of our knowledge, this represents the first approach for generating games based upon machine learned models of game design. 
Our major contributions are a new component-based representation of machine learned game design, an adaptation of conceptual expansion to this representation, and an evaluation demonstrating the ability to recreate existing games with this approach. 

\section{Related Work}

There exists a large amount of relevant prior work on automated game design. We can roughly separate this work into two major categories: (1) work that focuses on the generation of game stages, levels, and structure, and (2) work that focuses on the generation of game rules, mechanics, and dynamics. We note there exists many other components of games beyond structure and rules, such as visuals \cite{guzdial2017visual}, audio \cite{lopes2015sonancia}, and narrative \cite{li2014storytelling}. 

\subsection{Game Structure}

Much prior work in procedural content generation (PCG) has focused on generating game structure in the form of levels or puzzles for existing games \cite{hendrikx2013procedural}. PCG typically requires an authored, general level representation that can then be optimized by rules, constraints, and/or heuristics. There has been a great deal of PCG work for platformer levels, largely focused on Infinite Mario, a simplified Super Mario-like game \cite{karakovskiy2012mario}.

More recently, procedural content generation via machine learning (PCGML) has arisen as a means of training models to produce new game content given existing game content \cite{summerville2017procedural}. The benefit of PCGML is that it doesn't require expert design knowledge. Instead, it requires a dataset of existing content, and attempts to create more output like this input content, which limits novelty. Snodgrass and Ontan{\'o}n \shortcite{snodgrass2016approach} applied a transfer approach to adapt a model trained to generate levels for one game to another game. Guzdial and Riedl \shortcite{guzdial2016learning} made use concept blending \cite{fauconnier2001conceptual} to recombine machine learned models of Super Mario Bros. level design to produce new models capable of producing level types that do not exist in the original game.

\subsection{Game Rules}

Game rule generation has historically existed in the absence of generated structure, creating new rulesets for existing level designs. The majority of prior approaches to game rule generation have relied upon authoring a general game rule representation that is then constructed via grammars \cite{pell1992metagame}, optimized \cite{hom2007automatic,togelius2008experiment,browne2010evolutionary,cook2013mechanic} or constrained \cite{smith2010variations,zook2014automatic}.

The General Video Game Rule Generation track \cite{khalifa2017general} serves as a competition for ruleset generators in which generators are given a level and must create an appropriate ruleset. Khalifa et al. \shortcite{khalifa2017general} introduced two initial generators, a constructive approach based on human-authored rules and a genetic algorithm approach.

\subsection{Automated Game Design}

Treanor et al. \shortcite{treanor2012game} introduced Game-o-matic, a system for automatically designing games to match certain arguments or micro-rhetorics \cite{treanor2012micro}. This process created complete, if small, video games based on an authored procedure for transforming these arguments into games. Cook et al. produced the ANGELINA system for automated game design and development \cite{cook2017angelina}. There have been a variety of ANGELINA systems, each typically focusing on a particular game genre, using grammars and genetic algorithms to create game structure and rules \cite{cook2013mechanic}.

Nelson and Mateas \shortcite{nelson2008recombinable} introduce a system to swap rules in and out of game representations to create new experiences. Gow and Corneli \shortcite{gow2015towards} proposed applying conceptual blending to recombine games in the Video Game Description Language (VGDL) \cite{schaul2013video}. Nielsen et al. \shortcite{nielsen2015general} introduced an approach to mutate existing games expressed in the VGDL. Nelson et al. \shortcite{nelson2016mixed} defined a novel parameterized space of games and randomly alter subsets of parameters to explore the design space.

More recent work has applied variational autoencoders to produce noisy replications of existing games, called World Models or Neural Renderings \cite{ha2018world,eslami2018neural}. By its nature this work does not attempt to create new games. In addition, it is limited to only small subsections of existing games.

The approaches listed thus far in this section rely on creative recombinations or parameterization of designer-authored or crowdsourced \cite{guzdial2015crowdsourcing} representations of game knowledge. In this paper we focus on machine-learned game representations. Osborn et al. \shortcite{osborn2017automated} propose automated game design learning, an approach that makes use of emulated games to learn a representation of the game's structure \cite{osborn2017automatic} and rules \cite{summerville2017charda}. This approach is most similar to our own in terms of deriving a complete model of game structure and rules. However, this approach depends upon access to a game emulator and has no existing process for creating novel games.

\section{Approach}

Our approach to automated game design recombines machine learned representations of game design to produce novel games. The process is as follows: we take as input gameplay video and a spritesheet. A spritesheet is a collection of all of the images or sprites in the game, including all background art, animation frames, and components of level structure. We run image processing on the video with the spritesheet to determine where and what sprites occur in each frame. Then, we learn a model of level design and a ruleset for the game. We then merge the representations of level design and game ruleset into what we call a \emph{game graph}. A game graph is a graphical representation in which sprites become nodes and edges represent the level design and rules of a game. Finally, we apply our  {\em conceptual expansion} algorithm on these game graphs to produce a new graph from which novel games can be constructed.

At a high level, conceptual expansion creates a parameterized search space from an arbitrary number of input game graphs. One can think of each aspect of a game level design or ruleset design as a dimension in a high-dimensional space. Changing the value of a single dimension results in a new game design different from the original in some small way. A game graph provides a learned schematic for what dimensions exist and how they relate to each other.  With two or more game graphs, one can describe new games as interpolations between existing games, extrapolations along different dimensions, or alterations in complexity. The conceptual expansion algorithm searches this space to optimize for a particular goal or heuristic, thus creating new games. From these new game graphs we can reverse engineer level design models and rulesets to create new, playable games. In the following subsections we discuss the techniques we use to learn models of level design and game rulesets, how we create game graphs from the output of these techniques, and how we apply conceptual expansion to these game graphs.

\begin{figure}[tb]
	\centering
	\includegraphics[width=3in]{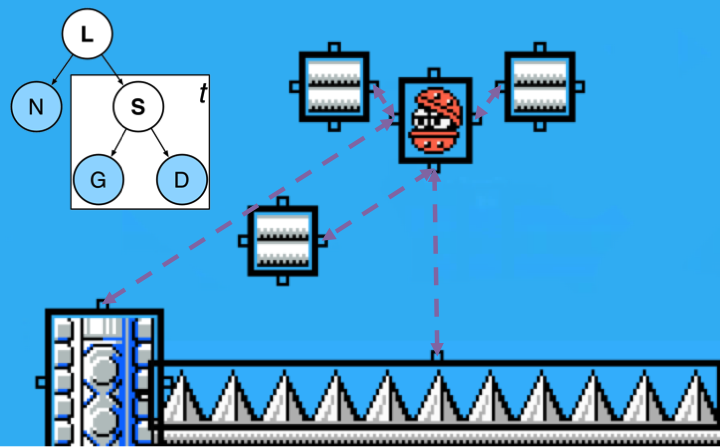}
	\caption{A visualization of the model (left) and basic building blocks from a subsection of the NES game Mega Man.}
	\label{fig:levelDesignModel}
\end{figure}

\subsection{Level Design Learning}

We use the technique from Guzdial and Riedl \shortcite{guzdial2016game} to learn a generative model of level design. At a high level this technique learns a hierarchical graphical model or bayesian network that represents probabilities over level structure. We visualize the abstract model and two base components in Figure \ref{fig:levelDesignModel}. There are a total of five types of nodes in the learned network: 

\begin{itemize}
\item \textbf{$G$: }Distribution over geometric shapes of sprite type $t$. In Figure \ref{fig:levelDesignModel} each box contains a G node value.
\item \textbf{$D$: }Distribution over relative positions of sprite type $t$. In Figure \ref{fig:levelDesignModel} all the purple lines represent a D node value.
\item \textbf{$N$: }Distribution of numbers of sprite types in a particular level chunk. For example in Figure \ref{fig:levelDesignModel} there are eleven spikes, three bars, one flying shell, etc.
\item \textbf{$S$: } The first hidden value, on which G and D nodes depend. S is the distribution of sprite styles for sprite type $t$, in terms of the distribution of geometric shapes and relative positions. That is, categories of sprites. For example, in Figure \ref{fig:levelDesignModel} there are three bars, but they all have the same S node value.
\item \textbf{$L$: }Distribution over level chunks.
\end{itemize}
\noindent
$L$ nodes are learned by a process of parsing gameplay video and iterative, hierarchical clustering. An ordering of $L$ nodes is learned to generate full levels. For further detail please see \cite{guzdial2016game}. Guzdial and Riedl \shortcite{guzdial2016game} also describe how to generate new level content from the model, and determine that the model's evaluation strongly correlates with human rankings of level design style.

\begin{figure}[tb]
	\centering
	\includegraphics[width=\columnwidth]{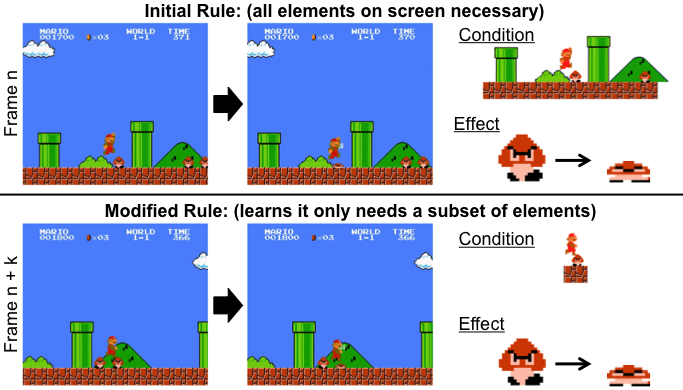}
	\caption{A visualization of two pairs of frames and an associated engine modification.}
	\label{fig:marioRuleExample}
\end{figure}

\subsection{Ruleset Learning}

We adapt the game engine learning work from Guzdial et al. \shortcite{guzdial2017game} to learn rulesets for individual games. This technique takes as input gameplay video and represents each frame as a list of conditional facts that are true in that frame. Whereas content modeling assumes the game is already known, this approach does not make this assumption. The fact types from the original paper are as follows:

\begin{itemize}
\item \textbf{$Animation$: } Each animation fact tracks a particular sprite seen in a frame by its name, width, and height. 
\item \textbf{$Spatial$: } Spatial facts track spatial information, the $x$ and $y$ locations of sprites on the screen.
\item \textbf{$RelationshipX/RelationshipY$: } The RelationshipX and RelationshipY facts track the relative positions of sprites to one another in their respective dimensions.
\item \textbf{$VelocityX/VelocityY$: }The VelocityX and VelocityY facts track the velocity of entities in their respective dimensions.
\item \textbf{$CameraX$: }Tracks the camera's $x$ position.
\end{itemize}
\noindent
The algorithm iterates through pairs of frames, using its current (initially empty) ruleset to attempt to predict the next frame. When a prediction fails it begins a process of iteratively updating the ruleset by adding, removing, and modifying rules to minimize the distance between the predicted and actual next frame. The rules are constructed with conditions and effects, where conditions are a set of facts that must exist in one frame and effects are a pair of facts where the second fact replaces the first when the rule fires. We visualize two examples of this process in Figure \ref{fig:marioRuleExample}. In the top row the difference of the goomba being squished or not is accounted for with a new goomba squishing rule that adds as a condition all facts from the first frame. Later when another pair of frames is encountered, that initial rule is modified to be more general. This occurs as the search heuristic prefers smaller, less complex rulesets. For further information on the ruleset learning process please see \cite{guzdial2017game}. The end result of this process is a sequence of rules that allows one to forward simulate the entire game state. In experiments, this learned ruleset allows a game playing agent to learn to play the game as well as the true ruleset.

We modify this technique by including two extra fact types. These are: 

\begin{itemize}
\item \textbf{$CameraY$: }Tracks the position of the camera in the $y$ dimension. We added this to cover games in which scrolling occurs in both $x$ and $y$ dimensions.
\item \textbf{$Random$: }Stand-in for a random variable (the original paper didn't account for random enemy behavior).
\end{itemize}
\noindent
We introduced the $Random$ fact as we parsed certain games in which enemy behavior depended on in-engine random variables. Thus, we added a possible modification to an existing rule to make it depend upon a random variable, which introduced a $Random$ fact to the conditions of a rule. If we did not include this, the final ruleset would include many rules where some random action (e.g. an enemy moving left or right) would be conditioned on extraneous detail in the frame (e.g. a relative distance to a background element) instead of a random variable.

\begin{figure}[tb]
	\centering
	\includegraphics[width=\columnwidth]{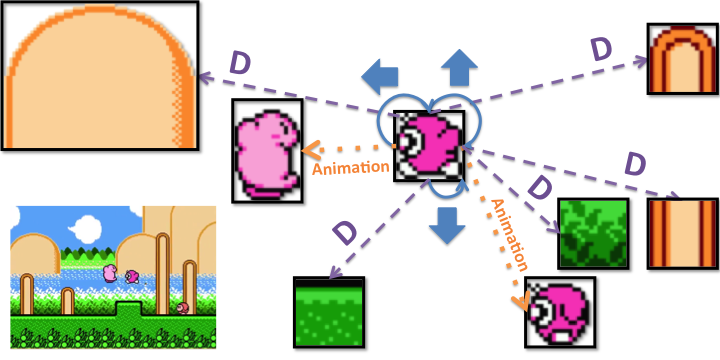}
	\caption{A subset of the game graph for one Waddle Doo enemy sprite from Kirby's Adventure and a relevant gameplay frame.}
	\label{fig:componentGraph}
\end{figure}

\subsection{Game Graph}

The output of the level design model learning process is a probabilistic graphical model. The output of the ruleset learning process is a sequence of formal-logic rules. We combine both of these into a single representation we call a \textit{game graph}. The construction of the game graph is straightforward. Each sprite in our original spritesheet becomes a node in an initially unconnected graph. Then, we add all the information from the level design model and ruleset representations as edges on this graph. There are five distinct types of edges with distinct types of values or weights:

\begin{itemize}
\item \textbf{G: } Stores the value of a $G$ node from the level design model, represented as the $x$ and $y$ positions of the shape of sprites, the shape of sprites (represented as a matrix), and a unique identifier for the $S$ and $L$ node that this $G$ node value depends on. This edge is cyclic, pointing to the same component it originated from. We note one might instead store this information as a value on the node itself, but treating it as an edge allows us to compare this and other cyclic edges to edges pointing between nodes.
\item \textbf{D: } Stores the value of a $D$ node connection, represented as a vector with the relative position, the probability, and a unique identifier for the $S$ and $L$ node that this $D$ node value depends on. This edge points to the equivalent node for the sprite this $D$ node connection connected to.
\item \textbf{N: } Stores the value of an $N$ node, which is a value representing a particular number of this component that can co-exist in the same level chunk and a unique Identifier for its $L$ node. This edge is cyclic.
\item \textbf{Rule condition: } Stores the value of a particular rule condition, which includes the information discussed for each fact type above and a unique identifier for the rule it is part of. This edge can be cyclic or can point to another component (as with Relationship facts).
\item \textbf{Rule effect: } Stores the value of a particular rule effect, which includes the information for both the pre and post facts and a unique identifier for the rule it is part of. This edge can be cyclic or can point to another component.
\end{itemize}

We visualize a small subsection of a final game graph for the Waddle Doo enemy from Kirby's Adventure in Figure \ref{fig:componentGraph}. The cyclic arrows in blue with arrows represent rule effects that impact velocity (we do not include the full rule for visibility). The orange dotted arrows represent rule effects that impact animation state. The dashed purple arrows represent $D$ node connection edges. We note that this is a very large graph, with dozens of nodes and thousands of edges. It contains all information from the learned level design model and game rulesets. That is, a game can be reconstructed from this graph.  Further, the graph structure can be manipulated in order to create new games by adding, deleting, or altering the values on edges.

\subsection{Conceptual Expansion over Game Graphs}

The size, complexity, and lack of uniformity of our game graph representation makes them ill-suited to generation approaches like machine learning or rule-based systems. Conceptual expansion is an approach from a class of algorithms that combine concepts \cite{guzdial2018creative}, extended to work on noisy, machine-learned knowledge representations. This makes it ideal for creating combinations of game graphs. We represent the conceptual expansion function as:
\begin{equation} \label{eq1}
  \mathrm{CE^X}(F,A) = a_1*f_1+a_2*f_2... a_n*f_n
\end{equation}

\noindent
Where $F=\{f_1...f_n\}$ is the set of all mapped features and $A=\{a_1, ... a_n\}$ is a filter representing what of and what amount of mapped feature $f_i$ should be represented in the final conceptual expansion. $X$ is the concept that we are attempting to represent with the conceptual expansion. The exact shape of $a_i$ depends upon the feature representation. If features are symbolic, $a_i$ can have values of either 0 or 1 (including the mapped feature or not), or vary from 0 to 1 if features are numeric or ordinal. Note that for numeric or ordinal values one may choose a different range (e.g. -1 to 1) dependent on the domain. In our case the $f$ values are individual nodes along with their incoming and outgoing edges, where $a$ is then a set of variables equal in size to the total number of the values on the incoming and outgoing edges. 
Thus, applying equation~\ref{eq1}, $CE^X$
is a final node in the new game graph, $f_1$, $f_2$... $f_n$ are existing nodes from our knowledge base, and $a_1$, $a_2$... $a_n$ are the filters on these nodes. 
A particular $a_i$ determines what edges from $f_i$ are added to the final expanded node $CE^{X}$ and how those edges are altered from the original edge in $f_i$. 
In this particular implementation we fix the number of final goal nodes, but this is not a requirement of the approach. Our new game graph is thus the set of conceptual expansions ($CE^{X}$) of each node ($X$) of that new graph. Essentially, each node of our new graph represents a combination of an arbitrary number of nodes from the existing graphs.

As an example, imagine that we do not have Goombas (the Mario enemy in Figure \ref{fig:marioRuleExample} as part of our knowledge base. A conceptual expansion for Goomba ($CE^{Goomba}$) might want to include many of the edges from the Waddle Doo node, such as the velocity rule effect to move left and fall ($f_{waddle doo1}$), but not the ability to jump ($f_{waddle doo2}$). One can encode this with a $a_{waddle doo}$ that filters out jump ($a_{waddle doo}$ = [1,0...]). We can alter $a_{waddle doo}$ to further specify we want the Goomba to move half as fast as the Waddle Doo. We might imagine other $f$ and corresponding $a$ values to incorporate other information from other game graph nodes for a final $CE^{Goomba}(F,A)$ that reasonably approximates a true Goomba node.

The conceptual expansion algorithm has two steps. First, a mapping is determined, which gives us the initial $a$ and $f$ values for each expanded node. This is accomplished by determining the best $n$ nodes in our knowledge base according to some heuristic $h$. For this paper we make use $n=10$ as an arbitrary starting point. We then fill in all the values of $a$ according to a normalized mapping in which the best mapped component has $a$ values of 1.0, and the $a$ values of the rest follow based upon their percentile. 

Given a mapping that determines the initial conceptual expansion, we optimize all of our expanded components according to the same heuristic $h$. For this implementation we make use of greedy hill-climbing, where we search for ten neighboring conceptual expansions and either select the best neighbor according to $h$ or return the current expansion if it has the best value in terms of $h$. To find a neighbor the system takes one of the following actions at random: (1) add a node to a particular component as an additional $f$ value with random $a$ values, (2) remove a random $f$ and $a$ pair, (3) for a particular expanded node change all values of $a$ by a random value [-2,2], and (4) alter a random value of a random $a$ value by multiplying it with a number uniformly chosen between [-2,2]. We chose [-2,2] as it the smallest whole number pair in which, with iterative alterations, we can get from any one value to any other value. 

\begin{table*}[tb]
\begin{center}
\begin{tabular}{ l|cc|cc|cc } 
  & \multicolumn{2}{c}{Mario} & \multicolumn{2}{c}{Kirby} & \multicolumn{2}{c}{Mega Man} \\ 
 \hline
  baseline & \specialcell{design error\\(train)} & \specialcell{rules error\\(test)} & \specialcell{design error\\(train)} & \specialcell{rules error\\(test)} & \specialcell{design error\\(train)} & \specialcell{rules error\\(test)} \\ 
 \hline
  Blend & 0.394 & 0.404 & 0.609 & 0.130 & 0.422 & 0.413 \\ 
 \hline
 KNN & 0.285  & 0.180 & 0.645  & 0.270 & 0.456 & \textbf{0.148} \\ 
 \hline
 GA & 0.285 & 0.180 & 0.645  & 0.134 & 0.454 & 0.163\\ 
 \hline
Expansion & \textbf{0.282} & \textbf{0.110} & \textbf{0.531} & \textbf{0.126}  & \textbf{0.396} &  0.189\\ 
 \hline
\end{tabular}
\caption{Errors for given level design evaluation. The systems were given a level design and tasked with creating a game with a matching level design and appropriate rules.}
\end{center}
\label{tab:trainLevel}
\end{table*}

\begin{table*}[tb]
\begin{center}
\begin{tabular}{ l|cc|cc|cc } 
  & \multicolumn{2}{c}{Mario} & \multicolumn{2}{c}{Kirby} & \multicolumn{2}{c}{Mega Man} \\ 
 \hline
  baseline & \specialcell{rules error\\(train)} & \specialcell{design error\\ (test)} & \specialcell{rules error\\(train)} & \specialcell{design error\\ (test)} & \specialcell{rules error\\(train)} & \specialcell{design error\\ (test)} \\ 
 \hline
  Blend & 0.152 & 0.623 &  0.09 & 0.664 & 0.127 & 0.727\\ 
 \hline
 KNN  & 0.180 & 0.400  & 0.105 & 0.638 & 0.148 & 0.525\\ 
 \hline
 GA & 0.143 & 0.537 & 0.103 & 0.672 & 0.143 & 0.592\\ 
 \hline
Expansion & \textbf{0.134} & \textbf{0.345}& \textbf{0.08} & \textbf{0.536}    & \textbf{0.118} & \textbf{0.502}\\ 
 \hline
\end{tabular}
\caption{Errors for the given rules evaluation. The systems were given a ruleset and tasked with creating a game with a matching ruleset and appropriate rules.}
\end{center}
\label{tab:trainEngine}
\end{table*}

\section{Evaluation}

Our ultimate goal for this automated game designer is for it to reduce the burden of game design and allow for the creation of novel games. This evaluation speaks to this goal, representing two hypothetical use-cases for a human designer and developer partner for the system. For evaluation purposes, we present a scenario where the system learns about two games and must try to construct a third, closely related game. To allow for comparative errors, the third game is a previously extant game that is unknown to the system. While this is not ideal, it serves to give an initial sense of the performance of this approach.

We began by parsing two gameplay videos of the first level of three games: (1) Super Mario Bros., (2) Kirby's Adventure, and (3) Mega Man. We chose these three games as they are platformers for the Nintendo Entertainment System and they vary in terms of mechanics and design. For example, Mega Man's levels move along the $y$-axis and include enemies with more complex behavior, impacting rules and level design. Kirby's Adventure devotes a fourth of its screen to the game's UI, which also impacts the rules and level design. We only parsed the first level to reduce the size of the final game graph, given the runtime of some of our baselines. 

For this evaluation we make use of a heuristic of distance from a partially specified goal game graph to some current game graph. We calculate the average of the distance from each component of the goal graph to the closest component in the current graph, normalized between [0,1]. For component to component distance we match the closest edges of the two components, where if two edges are of different types they will have a distance of 1.0, where otherwise the distance will vary from [0,1] depending on the number of matching values. We note that this heuristic is asymmetrical. This allows the system to measure fully defined game graphs in terms of partially defined game graphs. This distance function can be understood as an asymmetric Chamfer distance between game graphs.

We ran two distinct experiments. For the first we simulated the experience of a designer working with the system. We give the system game graphs for its initial two games, and the above heuristic with the goal of a third game graph with only the level design model knowledge. This represents the hypothetical situation in which a designer on some game wanted to create the full game but only had a level design defined. Our approach and the baselines then attempt to use feedback from the heuristic to create an entire game that matches both the goal level design model and completely unseen rules. We quantify the quality of the final game in terms of the distance from the indirectly accessible design model-only goal graph and the withheld rules-only graph. These can be understood as training and test error, respectively. For the second experiment we run the reverse, given only a rules-only game graph recreate an entire game that matches both rules and level design knowledge. For example, given Super Mario Bros. and Kirby's Adventure game graphs, and with a heuristic that gives indirect feedback on how close a candidate game graph is to the rules of Mega Man, to what extent can our approach and baselines recreate the entirety of Mega Man.

We constructed three baselines. They are as follows:
\begin{itemize}
\item \textbf{Blend: } We developed a conceptual blending version of our approach, given its history in other automated game design systems. This can be understood as a subset of the search space explored by conceptual expansions.
\item \textbf{KNN: } $K$-nearest neighbors represents one of the few machine learning approaches that can perform with such a small dataset. This approach returns whichever of the two game graphs is closest to the goal.
\item \textbf{GA: } We constructed a genetic algorithm baseline given its application to other automated game design systems. We make use of a population of size 10, initially made up of the original two game game graphs and four mutations of each. Our mutate function changes a random edge value to another value in the graph (e.g. sprite name to another sprite name, numeric value to another numeric value). Our crossover function creates a new graph by randomly selecting half of the nodes of each parent graph. We make use of the same heuristic/evaluation function as our system. This baseline was by far the slowest, taking between five and twenty times as long as our approach with a cap of one-hundred generations.
\end{itemize}


\section{Results}

We present the results for our designer-experiments (given level design heuristic, create level design and rules) in Table 1 and our developer-experiments in Table 2 (given rules heuristic, create rules and level design). Our approach outperformed the baselines in all but one case.

The developer-experiments appear significantly more challenging for all of the approaches. This matches our intuition that there could be many possible level designs for the same ruleset (e.g. there are multiple levels in a single game), while a particular level design model limits the kinds of rulesets that can successfully navigate the levels. For example, jump distances determine valid sizes of gaps. 

For the designer experiment with Mega Man as the goal the KNN baseline outperformed all approaches in terms of the rules (test) error. We anticipate the issue was that Mega Man's level design is not as tightly related to its rules, given that our approach and the GA baseline outperformed the KNN in terms of design (training) error. This follows from the fact that Mega Man receives a number of powerups that alter how he can move through the level.

\section{Discussion}

Our results demonstrate that our system can create games that more closely match a desired game given a distance to a partial specification of that game. Our system does not need a goal and could work with any appropriate heuristic. Thus one could create new games by optimizing for surprise, playability, or anything else one might consider.

For the sake of a quantitative evaluation, this paper focused on recreating existing games from a partial specification. However, the system can be used to create entirely novel games given a different heuristic. We anticipate that a full human subject study of generated games from our system will be required to investigate its value as a designer.

While the three games we chose as a dataset for this game differ from one another, they are all three platformer games originally published on the Nintendo Entertainment System, with two coming from the same company. Thus, we anticipate a need for a more complete study with multiple games from multiple eras of game design. In particular, we anticipate the need for new computer vision approaches for determining level geometry from video for 3D games and new fact types for games with significant mechanic variation (e.g. large amounts of menus, hidden information, etc.).

\section{Conclusions}

Automated game design has been restricted by the ability of its developers to find and encode high-quality design knowledge. We present an approach that generates new games by running conceptual expansion over a knowledge base of machine-learned representations of game designs. We evaluate our system in a simulated interaction with designers and developers on three classic games, and find that it out-performs state-of-the-art baselines on this task. To the best of our knowledge this represents the first machine learning-based automated game designer.

\section{Acknowledgements}

This material is based upon work supported by the National Science Foundation under Grant No. IIS-1525967. Any opinions, findings, and conclusions or recommendations expressed in this material are those of the author(s) and do
not necessarily reflect the views of the National Science Foundation.

\bibliographystyle{aaai}
\bibliography{aaai}

\begin{thebibliography}{}

\bibitem[\protect\citeauthoryear{Browne and
  Maire}{2010}]{browne2010evolutionary}
Browne, C., and Maire, F.
\newblock 2010.
\newblock Evolutionary game design.
\newblock {\em IEEE Transactions on Computational Intelligence and AI in Games}
  2(1):1--16.

\bibitem[\protect\citeauthoryear{Cook \bgroup et al\mbox.\egroup
  }{2013}]{cook2013mechanic}
Cook, M.; Colton, S.; Raad, A.; and Gow, J.
\newblock 2013.
\newblock Mechanic miner: Reflection-driven game mechanic discovery and level
  design.
\newblock In {\em European Conference on the Applications of Evolutionary
  Computation},  284--293.
\newblock Springer.

\bibitem[\protect\citeauthoryear{Cook, Colton, and
  Gow}{2017}]{cook2017angelina}
Cook, M.; Colton, S.; and Gow, J.
\newblock 2017.
\newblock The angelina videogame design system—part i.
\newblock {\em IEEE Transactions on Computational Intelligence and AI in Games}
  9(2):192--203.

\bibitem[\protect\citeauthoryear{Eslami \bgroup et al\mbox.\egroup
  }{2018}]{eslami2018neural}
Eslami, S.~A.; Rezende, D.~J.; Besse, F.; Viola, F.; Morcos, A.~S.; Garnelo,
  M.; Ruderman, A.; Rusu, A.~A.; Danihelka, I.; Gregor, K.; et~al.
\newblock 2018.
\newblock Neural scene representation and rendering.
\newblock {\em Science} 360(6394):1204--1210.

\bibitem[\protect\citeauthoryear{Fauconnier}{2001}]{fauconnier2001conceptual}
Fauconnier, G.
\newblock 2001.
\newblock Conceptual blending and analogy.
\newblock {\em The analogical mind: Perspectives from cognitive science}
  255--286.

\bibitem[\protect\citeauthoryear{Gow and Corneli}{2015}]{gow2015towards}
Gow, J., and Corneli, J.
\newblock 2015.
\newblock Towards generating novel games using conceptual blending.
\newblock In {\em Eleventh Artificial Intelligence and Interactive Digital
  Entertainment Conference}.

\bibitem[\protect\citeauthoryear{Guzdial and Riedl}{2016a}]{guzdial2016game}
Guzdial, M., and Riedl, M.
\newblock 2016a.
\newblock Game level generation from gameplay videos.
\newblock In {\em Twelfth Artificial Intelligence and Interactive Digital
  Entertainment Conference}.

\bibitem[\protect\citeauthoryear{Guzdial and
  Riedl}{2016b}]{guzdial2016learning}
Guzdial, M., and Riedl, M.
\newblock 2016b.
\newblock Learning to blend computer game levels.
\newblock In {\em Seventh International Conference on Computational
  Creativity}.

\bibitem[\protect\citeauthoryear{Guzdial \bgroup et al\mbox.\egroup
  }{2015}]{guzdial2015crowdsourcing}
Guzdial, M.; Harrison, B.; Li, B.; and Riedl, M.
\newblock 2015.
\newblock Crowdsourcing open interactive narrative.
\newblock In {\em Tenth International Conference on the Foundations of Digital
  Games}.

\bibitem[\protect\citeauthoryear{Guzdial \bgroup et al\mbox.\egroup
  }{2017}]{guzdial2017visual}
Guzdial, M.; Long, D.; Cassion, C.; and Das, A.
\newblock 2017.
\newblock Visual procedural content generation with an artificial abstract
  artist.
\newblock In {\em Third Computational Creativity and Games Workshop}.

\bibitem[\protect\citeauthoryear{Guzdial \bgroup et al\mbox.\egroup
  }{2018}]{guzdial2018creative}
Guzdial, M.; Liao, N.; Shah, V.; and Riedl, M.~O.
\newblock 2018.
\newblock Creative invention benchmark.
\newblock In {\em Ninth International Conference on Computational Creativity}.

\bibitem[\protect\citeauthoryear{Guzdial, Li, and
  Riedl}{2017}]{guzdial2017game}
Guzdial, M.; Li, B.; and Riedl, M.~O.
\newblock 2017.
\newblock Game engine learning from video.
\newblock In {\em 26th International Joint Conference on Artificial
  Intelligence}.

\bibitem[\protect\citeauthoryear{Ha and Schmidhuber}{2018}]{ha2018world}
Ha, D., and Schmidhuber, J.
\newblock 2018.
\newblock World models.
\newblock {\em arXiv preprint arXiv:1803.10122}.

\bibitem[\protect\citeauthoryear{Hendrikx \bgroup et al\mbox.\egroup
  }{2013}]{hendrikx2013procedural}
Hendrikx, M.; Meijer, S.; Van Der~Velden, J.; and Iosup, A.
\newblock 2013.
\newblock Procedural content generation for games: A survey.
\newblock {\em ACM Transactions on Multimedia Computing, Communications, and
  Applications (TOMM)} 9(1):1.

\bibitem[\protect\citeauthoryear{Hom and Marks}{2007}]{hom2007automatic}
Hom, V., and Marks, J.
\newblock 2007.
\newblock Automatic design of balanced board games.
\newblock In {\em Third Artificial Intelligence and Interactive Digital
  Entertainment},  25--30.

\bibitem[\protect\citeauthoryear{Karakovskiy and
  Togelius}{2012}]{karakovskiy2012mario}
Karakovskiy, S., and Togelius, J.
\newblock 2012.
\newblock The mario ai benchmark and competitions.
\newblock {\em IEEE Transactions on Computational Intelligence and AI in Games}
  4(1):55--67.

\bibitem[\protect\citeauthoryear{Khalifa \bgroup et al\mbox.\egroup
  }{2017}]{khalifa2017general}
Khalifa, A.; Green, M.~C.; Perez-Liebana, D.; and Togelius, J.
\newblock 2017.
\newblock General video game rule generation.
\newblock In {\em Computational Intelligence and Games}.

\bibitem[\protect\citeauthoryear{Li \bgroup et al\mbox.\egroup
  }{2014}]{li2014storytelling}
Li, B.; Thakkar, M.; Wang, Y.; and Riedl, M.~O.
\newblock 2014.
\newblock Storytelling with adjustable narrator styles and sentiments.
\newblock In {\em International Conference on Interactive Digital
  Storytelling},  1--12.
\newblock Springer.

\bibitem[\protect\citeauthoryear{Lopes, Liapis, and
  Yannakakis}{2015}]{lopes2015sonancia}
Lopes, P.; Liapis, A.; and Yannakakis, G.~N.
\newblock 2015.
\newblock Sonancia: Sonification of procedurally generated game levels.
\newblock In {\em First Computational Creativity and Games Workshop}.

\bibitem[\protect\citeauthoryear{Nelson and
  Mateas}{2008}]{nelson2008recombinable}
Nelson, M.~J., and Mateas, M.
\newblock 2008.
\newblock Recombinable game mechanics for automated design support.
\newblock In {\em Fourth Artificial Intelligence and Interactive Digital
  Entertainment Conference}.

\bibitem[\protect\citeauthoryear{Nelson \bgroup et al\mbox.\egroup
  }{2016}]{nelson2016mixed}
Nelson, M.; Colton, S.; Powley, E.; Gaudl, S.; Ivey, P.; Saunders, R.;
  Perez~Ferrer, B.; and Cook, M.
\newblock 2016.
\newblock Mixed-initiative approaches to on-device mobile game design.
\newblock In {\em CHI Workshop on Mixed-Initiative Creative Interfaces}.

\bibitem[\protect\citeauthoryear{Nielsen \bgroup et al\mbox.\egroup
  }{2015}]{nielsen2015general}
Nielsen, T.~S.; Barros, G.~A.; Togelius, J.; and Nelson, M.~J.
\newblock 2015.
\newblock General video game evaluation using relative algorithm performance
  profiles.
\newblock In {\em European Conference on the Applications of Evolutionary
  Computation},  369--380.
\newblock Springer.

\bibitem[\protect\citeauthoryear{Osborn, Summerville, and
  Mateas}{2017a}]{osborn2017automatic}
Osborn, J.; Summerville, A.; and Mateas, M.
\newblock 2017a.
\newblock Automatic mapping of nes games with mappy.
\newblock In {\em Twelfth International Conference on the Foundations of
  Digital Games}, ~78.
\newblock ACM.

\bibitem[\protect\citeauthoryear{Osborn, Summerville, and
  Mateas}{2017b}]{osborn2017automated}
Osborn, J.~C.; Summerville, A.; and Mateas, M.
\newblock 2017b.
\newblock Automated game design learning.
\newblock In {\em Computational Intelligence and Games},  240--247.

\bibitem[\protect\citeauthoryear{Pell}{1992}]{pell1992metagame}
Pell, B.
\newblock 1992.
\newblock Metagame in symmetric chess-like games.

\bibitem[\protect\citeauthoryear{Schaul}{2013}]{schaul2013video}
Schaul, T.
\newblock 2013.
\newblock A video game description language for model-based or interactive
  learning.
\newblock In {\em Computational Intelligence in Games},  1--8.

\bibitem[\protect\citeauthoryear{Smith and Mateas}{2010}]{smith2010variations}
Smith, A.~M., and Mateas, M.
\newblock 2010.
\newblock Variations forever: Flexibly generating rulesets from a sculptable
  design space of mini-games.
\newblock In {\em Computational Intelligence and Games},  273--280.

\bibitem[\protect\citeauthoryear{Snodgrass and
  Ontan{\'o}n}{2016}]{snodgrass2016approach}
Snodgrass, S., and Ontan{\'o}n, S.
\newblock 2016.
\newblock An approach to domain transfer in procedural content generation of
  two-dimensional videogame levels.
\newblock In {\em Twelfth Artificial Intelligence and Interactive Digital
  Entertainment Conference}.

\bibitem[\protect\citeauthoryear{Summerville \bgroup et al\mbox.\egroup
  }{2017}]{summerville2017procedural}
Summerville, A.; Snodgrass, S.; Guzdial, M.; Holmg{\aa}rd, C.; Hoover, A.~K.;
  Isaksen, A.; Nealen, A.; and Togelius, J.
\newblock 2017.
\newblock Procedural content generation via machine learning (pcgml).
\newblock {\em arXiv preprint arXiv:1702.00539}.

\bibitem[\protect\citeauthoryear{Summerville, Osborn, and
  Mateas}{2017}]{summerville2017charda}
Summerville, A.; Osborn, J.; and Mateas, M.
\newblock 2017.
\newblock Charda: Causal hybrid automata recovery via dynamic analysis.
\newblock {\em arXiv preprint arXiv:1707.03336}.

\bibitem[\protect\citeauthoryear{Togelius and
  Schmidhuber}{2008}]{togelius2008experiment}
Togelius, J., and Schmidhuber, J.
\newblock 2008.
\newblock An experiment in automatic game design.
\newblock In {\em Computational Intelligence and Games},  111--118.

\bibitem[\protect\citeauthoryear{Treanor \bgroup et al\mbox.\egroup
  }{2012a}]{treanor2012game}
Treanor, M.; Blackford, B.; Mateas, M.; and Bogost, I.
\newblock 2012a.
\newblock Game-o-matic: Generating videogames that represent ideas.
\newblock In {\em Third workshop on Procedural Content Generation in Games}.

\bibitem[\protect\citeauthoryear{Treanor \bgroup et al\mbox.\egroup
  }{2012b}]{treanor2012micro}
Treanor, M.; Schweizer, B.; Bogost, I.; and Mateas, M.
\newblock 2012b.
\newblock The micro-rhetorics of game-o-matic.
\newblock In {\em Seventh International Conference on the Foundations of
  Digital Games}.

\bibitem[\protect\citeauthoryear{Zook and Riedl}{2014}]{zook2014automatic}
Zook, A., and Riedl, M.~O.
\newblock 2014.
\newblock Automatic game design via mechanic generation.
\newblock In {\em AAAI},  530--537.

\end{thebibliography}

\end{document}